\documentclass[conference]{IEEEtran}
\IEEEoverridecommandlockouts
\usepackage{cite}
\usepackage{amsmath,amssymb,amsfonts}
\usepackage{algorithmic}
\usepackage{graphicx}
\usepackage{textcomp}
\usepackage{xcolor}
\usepackage[hyphens]{url}
\usepackage{hyperref}
\usepackage[hyphenbreaks]{breakurl}

\usepackage[nolist]{acronym}

\newcommand{\snn}[1]{\hyperref[snn:#1]{P#1}}
\usepackage{tikz}
\usepackage{tabularx}

\def\BibTeX{{\rm B\kern-.05em{\sc i\kern-.025em b}\kern-.08em
    T\kern-.1667em\lower.7ex\hbox{E}\kern-.125emX}}
\begin{document}

\title{Low-Power Vibration-Based Predictive Maintenance for Industry 4.0 using Neural Networks: A Survey
}

\author{\IEEEauthorblockN{Alexandru Vasilache,\!$^{1,2}$ Sven Nitzsche,\!$^{1,2}$ Daniel Floegel, \!$^{1,2}$ Tobias Schuermann, \!$^{1,2}$ Stefan von Dosky,\!$^{3}$ \\ 
Thomas Bierweiler,\!$^{3}$ Marvin Mu\ss ler,\!$^{3}$ Florian K\"alber,\!$^{4}$ Soeren Hohmann,\!$^{2}$ Juergen Becker\!$^{2}$}
    \IEEEauthorblockA{\textit{$^{1}$ FZI Research Center for Information Technology, Karlsruhe, Germany} \\
    \textit{$^{2}$ Karlsruhe Institute of Technology, Karlsruhe, Germany}\\
    \textit{$^{3}$ Siemens, Digital Industries, Process Automation, Karlsruhe, Germany}\\
    \textit{$^{4}$ NXP Semiconductors Germany GmbH, Munich, Germany} \\
    }
}

\maketitle
\begin{acronym}[Longest Abrev] 
\acro{PM}{Predictive Maintenance}
\acro{ANN}{Artificial Neuronal Network}
\acro{AE}{Auto-Encoder}
\acro{DBN}{Deep Belief Network}
\acro{CNN}{Convolutional Neural Network}
\acro{DNN}{Deep Neural Network}
\acro{LIF}[LIF]{Leaky Integrate-and-Fire}
\acro{IF}[IF]{Integrate-and-Fire}
\acro{LI}[LI]{Leaky Integrate}
\acro{ALIF}{Adaptive Leaky Integrate-and-Fire}
\acro{adex}[AdEx]{adaptive exponential integrate-and-fire}
\acro{LSTM}{Long Short-Term Memory}
\acro{RNN}{Reccurent Neural Network}
\acro{LSNN}{Long Short-Term Spiking Neural Network}
\acro{NN}{Neural Network}
\acro{SNN}{Spiking Neural Network}
\acro{ML}{Machine Learning}
\acro{DL}{Deep Learning}
\acro{CWRU}{Case Western Reserve University}
\acro{AI}{artificial intelligence}
\acro{KPI}{Key Performance Indicator}
\acro{RUL}{remaining useful life}
\acro{soc}[SoC]{system-on-chip}
\acro{BLE}{Bluetooth Low Energy}
\acro{IBF}{Induced Bearing Fault}
\acro{PUD}{Paderborn University Dataset}
\acro{rtf}[R2F]{Run-To-Failure Bearing Fault}
\acro{SHM}{Structural Health Monitoring}
\acro{DFT}{Discrete Fourier Transform}
\acro{FFT}{Fast Fourier Transform}
\acro{STFT}{Short-Time Fourier transform}
\acro{BPTT}{Backpropagation Through Time}
\acro{BP}{Backpropagation}
\acro{MCC}{Matthews Correlation Coefficient}
\acro{LMD}{Local Mean Decomposition}
\acro{GRF}{Gaussian Receptive Field}
\acro{TTFS}{Time-to-First-Spike}
\acro{PCA}{Principal component analysis}
\acro{PSRM}{Probabilistic Spiking Response Model}
\acro{SRM}{Spiking Response Model}
\acro{SLT}{Slantlet Transform}
\acro{GFB}{Gammatone filter bank}

\end{acronym}

\begin{abstract}

The advancements in smart sensors for Industry 4.0 offer ample opportunities for low-powered predictive maintenance and condition monitoring. 
However, traditional approaches in this field rely on processing in the cloud, which incurs high costs in energy and storage. 
This paper investigates the potential of neural networks for low-power on-device computation of vibration sensor data for predictive maintenance.
We review the literature on \acp{SNN} and \acp{ANN} for vibration-based predictive maintenance by analyzing datasets, data preprocessing, network architectures, and hardware implementations.
Our findings suggest that no satisfactory standard benchmark dataset exists for evaluating neural networks in predictive maintenance tasks. Furthermore frequency domain transformations are commonly employed for preprocessing. 
\acp{SNN} mainly use shallow feed forward architectures, whereas \acp{ANN} explore a wider range of models and deeper networks.
Finally, we highlight the need for future research on hardware implementations of neural networks for low-power predictive maintenance applications and the development of a standardized benchmark dataset.

\end{abstract}

\begin{IEEEkeywords}
predictive maintenance (PM), low-power, vibration-based condition monitoring, Industry 4.0, neural networks (NNs), spiking neural networks (SNNs), artificial neural networks (ANNs), edge computing, on-device processing
\end{IEEEkeywords}

\section{Introduction}
Over the past years, smart sensors have been on the rise in the industry to collect data from previously uninstrumented components \cite{noauthor_smart_nodate}. In addition to simple data such as temperature and humidity, acoustic data (structure-borne sound and air-borne sound) are of particular interest, as they often represent the technical condition of a component \cite{vishwakarma2017vibration} \cite{jombo2023acoustic}. Using this data for \ac{PM} can reduce upkeep costs and increase production rates. Traditional approaches to \ac{PM} generally employ a central processing instance (cloud) for analyzing the data and computing the \acp{KPI}. However, using a cloud instance necessitates transporting and storing high-sampled data, resulting in high energy and storage costs.

Since the energy required for data transmission is potentially more significant than the energy required for data processing, the lifetime of the sensor can increase by improving the transmission energy consumption. More energy-efficient transmission can be achieved by employing modern wireless, narrowband, and cost-effective communication channels such as the Narrowband Internet of Things (NB-IoT). Further improvements can be gained by reducing the overall amount of transmitted data. Moving the data processing from the cloud to the sensor can thus reduce the transmission requirements to only the \acp{KPI}.

\section{Focused Applications} 
The structure-born vibrations of rotating machinery can offer meaningful information about its state of health.
Those vibration signals can be utilized to identify anomalies, levels of wear and tear and other signs of faults of machines like pumps, compressors or industrial drives.
For motors, the bearing condition, misalignment, and asymmetrical winding forces, e.g., due to a local short circuit, can be monitored to avoid unsuspected breakdowns or indicate the remaining useful life.
Non-stationary operating machines and low rotational speeds yield vibration signals which pose a major challenge for \ac{PM}.

Airborne vibrations (sound) can provide further information about the condition of machines.
This approach is sensible due to the non-contact measurement and the significantly larger acoustic bandwidth.
A typical disadvantage is the superimposition of ambient noise. 
However, depending on the number of microphones and the signal processing, it is possible to detect and localize compressed gas leaks, for example. 
Intelligent engine sensors equipped with MEMS microphones that keep the noise emission of an engine below a standardized limit for the health protection of personnel provide another example of the usefulness of airborne sound in \ac{PM}. 
Other applications that benefit from the non-contact nature of sensor technology include the indication of extraneous discharges on high-voltage transformers and the condition monitoring of turbochargers driven by hot exhaust gases.

\section{Sensors}
The proliferation of mobile devices has significantly accelerated the growth of MEMS sensors, such as microphones, accelerometers, compasses, pressure sensors, light sensors, and capacitive touchscreens. Modern microcontrollers integrate these sensors with fast digital and wireless interfaces, simplifying device development and maintenance. Tiny boards with multiple MEMS sensors are now widely available, enhancing consumer products and other applications \cite{1noauthor_stmicroelectronics_nodate}. The adoption of \ac{BLE} wireless interfaces has further boosted this trend due to its high compatibility and market acceptance. Potential counterparts for sensor \acp{soc} include laptops, tablets, and smartphones, all of which benefit from the low peak current demand. Energy-efficient \ac{BLE} \acp{soc} now feature ultra-low power radio transceivers \cite{2noauthor_efr32bg27_nodate} and long-range modes with bit coding, along with vendor-maintained security libraries. Industry vendors have also contributed to the development and early trials of these systems, often supported by smartphone apps and over-the-air firmware updates, enabling fully remote sensor operation with a gateway and cloud integration \cite{3noauthor_tidc-cc2650stk-sensortag_nodate}.
MEMS sensor performance has also advanced. For example, the \cite{4noauthor_lsm6dsl_nodate} sensor's three-axis readout rate of 6664 Hz is well-suited for industrial applications like pump monitoring based on structure-borne noise. Its vibration transfer function remains relatively flat up to about 1000 Hz, offering good sensitivity and 16-bit sample resolution, regardless of the measurement range. MEMS digital microphones with extended frequency ranges, including ultrasound, are particularly valuable for detecting early-stage damage, such as in ball bearings \cite{5noauthor_new_nodate}.
In addition to countless small companies, several established process sensor suppliers \cite{6noauthor_sitrans_nodate, 7noauthor_condition_nodate, 8noauthor_honeywell_nodate, 9corporation_xs770a_2019} offer products and systems that include sensors, gateways, and cloud applications.

\section{Spiking Neural Networks}
\label{sec:SNN}

\subsection{Fundamentals}
\label{sec-snn-basics}
Spiking neural networks are a biologically inspired type of neural network, which uses discrete impulses to transport information. As spikes are typically unary, with no specific value attached to them, the actual information is encoded by the timing between spikes. When there is no spike entering a neuron at a specific point in time, this neuron stays inactive and no computation is required. This makes \acp{SNN} a promising candidate for deploying AI to embedded systems.

\subsection{Preprocessing / Event Conversion}
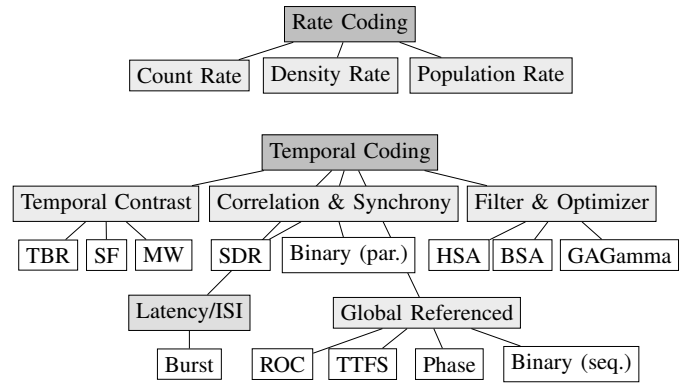
\begin{figure}[]
  \centering
\resizebox{1\linewidth}{!}{%
\begin{tikzpicture}[
    scale=1,
    node distance=1.5cm,
    every node/.style={draw, rectangle, fill=gray!0},
    level distance=2cm
]
    \node (rate_coding) at (0,0) [fill=gray!50] {Rate Coding}
        child {node (count_rate) at (-1,1.2) [fill=gray!15] {Count Rate}}
        child {node (density_rate) at (-0.3,1.2) [fill=gray!15] {Density Rate}}
        child {node (population_rate) at (0.7,1.2) [fill=gray!15] {Population Rate}};

    \node (temporal_coding) at (0,-2) [fill=gray!50] {Temporal Coding}
        child {node (temporal_contrast) at (-0.8,1.2) [fill=gray!15] {Temporal Contrast}
            child {node (TBR) at (0.6,1.2) {TBR}}
            child {node (SF) at (0.01,1.2) {SF}}
            child {node (MW) at (-0.6,1.2) {MW}}
        }
        child {node (latency_isi) at (-1,-0.5) [fill=gray!25] {Latency/ISI}
            child {node (burst) at (0,1.2) {Burst}}
        }
        child {node (global_referenced) at (1.2,-0.5) [fill=gray!15] {Global Referenced}
                child {node (ROC) at (0,1.2) {ROC}}
                child {node (TTFS) at (-0.25,1.2) {TTFS}}
                child {node (phase) at (-0.4,1.2) {Phase}}
                child {node (binary_seq) at (0,1.2) {Binary (seq.)}}
        }
        child {node (correlation_synchrony) at (-1.75,1.2) [fill=gray!15] {Correlation \& Synchrony}
            child {node (SDR) at (-0.7,1.2) {SDR}}
            child {node (binary_par) at (-0.52,1.2) {Binary (par.)}}
        }
        child {node (filter_optimizer) at (0.3,1.2) [fill=gray!15] {Filter \& Optimizer}
            child {node (HSA) at (-0.1,1.2) {HSA}}
            child {node (BSA) at (-0.61,1.2) {BSA}}
            child {node (GAGamma) at (-0.6,1.2) {GAGamma}}                
        };
\end{tikzpicture}}
  \caption{Taxonomy of spike encoding techniques. Inspired by \cite{auge2021survey}.}
  \label{fig:spike_enc_overview}
 \end{figure}

A major challenge when using \acp{SNN} for vibration-based \ac{PM} is the lack of event-based sensors.
As described in Section~\ref{sec-snn-basics}, \acp{SNN} commonly require spikes as inputs rather than real-valued data. These spikes, outside of an \ac{SNN} typically also referred to as events, should be retrieved directly from an event-based sensor for maximum efficiency. However, until now only event-based cameras have majored enough to be available as off-the-shelf components that could be used for fully event-based predictive maintenance~\cite{li_ebc_pm_2024}. Other sensors, such as event-based audio sensors are actively researched~\cite{lenk2023neuromorphic}, but not yet freely available. Alternatively, events can be created by converting conventional sensor data. For this purpose, various approaches exist. They can be roughly divided into rate-based and temporal coding schemes, as shown in Fig.~\ref{fig:spike_enc_overview}.
In general, temporal coding schemes use less spikes to encode information than rate-based approaches. Less spikes require less calculation steps of the \ac{SNN} used to process the event data and hence are beneficial for high energy efficiency. 
Auge et al. give an overview of available conversion approaches~\cite{auge2021survey}. They describe the biophysical background of spike encoding and review implementations of various schemes. The specific encoding technique to use is dependent on the actual use case~\cite{schuman2019non, schuman2022evaluating}. However, there is no clear indication of which technique might be beneficial for vibration-based predictive maintenance in the state-of-the-art.

\subsection{Approaches}

The following subsection will present the most relevant approaches to Low-Power Vibration-Based Predictive Maintenance for Industry 4.0 that use Spiking Neural Networks (SNNs).

\label{snn:1}
\subsubsection{Ultra-low Power Machinery Fault Detection Using Deep Neural Networks\cite{nitzsche_ultra-low_2021}}
The authors present a work-in-progress approach to vibration-based fault detection in centrifugal pumps (\ac{CWRU} dataset \cite{loparo2012case}) and bearing faults (\ac{PUD} \cite{noauthor_konstruktions_nodate}). The vibration data is fed into the \ac{SNN} by use of Current Injection. They employ a supervised approach to training a 4 layer \ac{CNN} (input, output, two hidden layers), then convert it to a rate-coded \ac{SNN} with \ac{LIF} neurons using NengoDL \cite{bekolay2014nengo}.

\label{snn:2}
\subsubsection{Online Detection of Vibration Anomalies Using Balanced Spiking Neural Networks \cite{dennler_online_2021}}
This approach uses balanced spiking neural networks (BSNNs) inspired by Efficient Balance Networks \cite{bourdoukan2012learning} for unsupervised, online anomaly detection using vibration sensor data from bearing faults. The authors present an implementation of the networks on the Brian2 framework \cite{stimberg2019brian} and the DYNAP-SE chip \cite{moradi2017scalable}. The data used includes the \ac{IBF} dataset \cite{bechhoefer2013condition} and \ac{rtf} dataset \cite{qiu2006wavelet}. The preprocessing pipeline includes frequency decomposition using Gammatone filter banks \cite{patterson_efficient_1987} based on the Cochlea model and an asynchronous delta modulator \cite{corradi2015neuromorphic} \cite{sharifshazileh2021electronic} to convert the continuous values to spikes. The model consists of 3 layers (input layer, one hidden \ac{LIF} layer, output layer with one output neuron). Results show a perfect confusion matrix for \ac{IBF}, which detects all healthy transitions, while \ac{rtf} shows earlier anomaly detection compared to state-of-the-art methods for half of the dataset.

\label{snn:3}
\subsubsection{Damage Detection in Structural Health Monitoring with Spiking Neural Networks \cite{zanatta_damage_2021}}
This study focuses on using \acp{LSNN} with \ac{ALIF} neurons for supervised damage detection tasks. The architecture consists of an input layer, connected to two recurrent ensembles, which are connected to the output layer. The study utilizes vibration sensor data from a \ac{SHM} dataset obtained from a bridge subjected to \ac{SHM}. The preprocessing stage consists of extracting spectral features by use of the \ac{DFT}, which are then fed to the \ac{SNN} by current injection. The training method uses an approximation of \ac{BPTT} called E-prop \cite{bellec2020solution}. Comparisons with alternative \ac{ANN} models show that \acp{LSNN} have similar accuracy. The \ac{LSNN} that performed the best achieved a \ac{MCC} of 0.88 when distinguishing between damaged and healthy bridge conditions.
    
\label{snn:4}
\subsubsection{Spiking Neural Network-Based Near-Sensor Computing for Damage Detection in Structural Health Monitoring \cite{barchi_spiking_2021}}
This study presents an approach to \ac{SHM} using \acp{LSNN} with 3 layers (one input, one output, one hidden recurrent with 20 \ac{ALIF} neurons) trained in a supervised manner with \ac{BPTT} \cite{bellec2020solution}. The study further presents an implementation of the model on an STM32F407VG6 MCU \ac{soc} \cite{noauthor_stm32f407vg_nodate}. The paper utilizes vibration data from a \ac{SHM} dataset obtained from a viaduct that underwent an intervention to strengthen its structure. The preprocessing involves spectral coefficient analysis of accelerometer waveforms, followed by an exploration of two training methods: Current-based, where \ac{FFT} coefficients are applied as a constant current to input neurons, and Event-based, where \ac{FFT} coefficients are transformed to spikes using a \ac{TTFS} based method. The best results show a \ac{MCC} $\geq$ 0.75.

\label{snn:5}
\subsubsection{A spiking neural network-based approach to bearing fault diagnosis \cite{zuo_spiking_2021}}
This paper proposes a \ac{SNN}-based methodology for supervised bearing fault diagnosis. It explores the use of \acp{SNN} with vibration sensor data obtained from bearing fault datasets, specifically the \ac{CWRU} dataset \cite{loparo2012case} and the MFPT dataset \cite{noauthor_fault_nodate}. The preprocessing includes \ac{LMD} combined with population coding using \acp{GRF} \cite{yu2014brain} and \ac{TTFS} coding. The model training uses an improved tempotron learning rule \cite{gutig2006tempotron}, which optimizes synaptic weights by minimizing the potential difference between the firing threshold and actual membrane potential. The \ac{SNN} architecture consists of two \ac{LIF} neuron layers (one input, one output, no hidden). Results show high accuracy, with \ac{CWRU} reaching 99.17\% and MFPT reaching 99.54\%.

\label{snn:6}
\subsubsection{Research on Fault Diagnosis Based on Spiking Neural Networks in Deep Space Environment \cite{li_research_2022}}
The study deals with fault diagnosis in deep space environments using \acp{SNN} in a supervised framework. The main dataset explored throughout the paper is the full life cycle data of the deep space detector bearing in the NASA database, with verification of the method being performed on the \ac{CWRU} dataset \cite{loparo2012case} as well. The bearing fault data is analyzed for fault diagnosis with vibration sensors as the primary input. Preprocessing includes the time-frequency domain \ac{LMD}, min-max normalization, and Gaussian population coding. Training involves the supervised training of \acp{ANN} and their conversion into \acp{SNN}. The model consists of one hidden layer with \ac{LIF} neurons, one input and one output layer. Results demonstrate the efficiency of SNNs compared to traditional \ac{CNN} models, showing shorter training times and high accuracy on various datasets.

\label{snn:7}
\subsubsection{Machine Hearing for Industrial Acoustic Monitoring using Cochleagram and Spiking Neural Network \cite{zhang_machine_2022}}
This paper investigates the application of machine hearing to industrial acoustic monitoring using \acp{SNN} for fault diagnosis, with a focusing on bearing fault detection. Data is obtained by taking 10 second acoustic measurements of a GUNT PT500 Machinery Diagnostic System \cite{noauthor_products_nodate} with each of the 6 bearing fault conditions (a normal bearing condition, a bearing condition with an outer race defect, an inner race defect, a roller element defect, and combined damages, and a bearing condition that is severely worn). The pre-processing involves the Cochleagram, which models the frequency filtering characteristics of the cochlea through Gammatone filters \cite{patterson_efficient_1987}, combined with \ac{PCA} to reduce the number of frequency features from 128 to 50. The data is then encoded into spikes with population coding using neurons with \acp{GRF} and \ac{TTFS} encoding. The \ac{SNN} model consists of 3 layers (one input, one output, one hidden), employing a threshold-based neuron model. Training uses margin maximization techniques \cite{dora2021spiking}. Results show a classification accuracy of 89.66\% in detecting the bearing fault states, which is comparable to alternative techniques such as \acp{RNN} with two \ac{LSTM} layers (94.7\%).

\label{snn:8}
\subsubsection{Efficient Time Series Classification using Spiking Reservoir \cite{dey_efficient_2022}}
This paper focuses on reservoir \acp{SNN} for fault detection applications. Using vibration data, the authors examine real-time classification datasets from the UCR repository \cite{dau2019ucr}, including two engine noise datasets, an inline process control measurement dataset, and a seismic dataset. Their approach compares the effectiveness of Poisson rate coding versus Gaussian temporal coding for preprocessing. The \ac{LIF}-based model consists of a reservoir of 2 layers of neurons (excitatory and inhibitory). Training is performed using a supervised approach, by feeding the neuronal trace values of the excitatory neurons into a Logistic Regression based classifier that is trained with the corresponding class labels. The results show that the Gaussian temporal coding yields superior accuracy to the Poisson rate coding. Furthermore, the Gaussian temporal encoding network outperforms the IAL-Edge \cite{pal2020instant} comparison in all datasets except seismic while exhibiting performance close to that of the TN-C comparison \cite{malhotra2017timenet}.

\label{snn:9}
\subsubsection{A multi-layer spiking neural network-based approach to bearing fault diagnosis \cite{zuo_multi-layer_2022}}
The study investigates multilayer \acp{SNN}. It focuses on bearing fault diagnosis of vibration data from three bearing databases, including the MFPT \cite{noauthor_fault_nodate}, the \ac{CWRU} dataset \cite{loparo2012case}, and the \ac{PUD} \cite{noauthor_konstruktions_nodate}. Preprocessing includes \ac{LMD}, \acp{GRF}, and Boltzmann distribution-based pulse probability sequence conversion. The model consists of a multilayer \ac{SNN} with a \ac{PSRM} neuron model and was trained using \ac{BP}. This approach outperformed existing methods regarding accuracy on all three datasets.

\label{snn:10}
\subsubsection{Novel Spiking Neural Network Model for Gear Fault Diagnosis \cite{ali_novel_2022}}
In this paper, a \ac{SNN} model with 2 layers (one input, one output) focused on gear fault diagnosis has been developed, by use of \ac{SRM} neurons. The dataset used for training and testing consists of self-recorded acoustic gear fault data, with one healthy and five faulty classes. The data was pre-processed using the \ac{SLT} to transform the time- into the frequency domain. Sixteen features were extracted, including 11 time-based features and five frequency-based features. The results showed a diagnostic accuracy of 95\%.

\label{snn:11}
\subsubsection{Spiking Neural Networks for Structural Health Monitoring \cite{joseph_spiking_2022}}
This paper explores the application of \acp{SNN} for \ac{SHM}, focusing on unsupervised anomaly detection using vibration sensor data. The study uses a simulated \ac{SHM} dataset generated by exciting a single-degree-of-freedom linear oscillator with Gaussian white noise forcing. Preprocessing involves using 36 \acp{GFB} \cite{patterson_efficient_1987} to capture frequency characteristics, followed by spike encoding using current injection. The training uses a Neural Engineering Framework (NEF) \cite{eliasmith2003neural} to map cepstrum features to the \ac{SNN} implementation by Nengo \cite{bekolay2014nengo}. The model consists of \ac{LIF} neurons arranged in 2 layers (one input and one output layer). Results show that an averaging window of 1.5 seconds yields a clear separation between damaged and undamaged states.

\label{snn:12}
\subsubsection{Comparing Reservoir Artificial and Spiking Neural Networks in Machine Fault Detection Tasks \cite{kholkin_comparing_2023}}
This paper investigates and compares the performance of reservoir \acp{ANN} and \acp{SNN} in supervised fault detection tasks using vibration sensor data of bearing and gearbox faults. The datasets used include the ETU Bearing Dataset, the 
\ac{CWRU} dataset \cite{loparo2012case}, and a Kaggle Gearbox Fault Diagnosis Dataset \cite{noauthor_gearbox_nodate}. Preprocessing includes spectral analysis with \ac{STFT} feature extraction. The models include Echo State Networks (ESNs) \cite{gallicchio2013tree} and Liquid State Machines (LSMs) \cite{maass2002real} with \ac{adex} neurons \cite{brette2005adaptive} arranged in three layers (input, one hidden reservoir, output). Despite longer execution times, \acp{SNN} show superior classification accuracy compared to \acp{ANN}.

\begin{table*}
\centering
\caption{\ac{SNN} Features in Reviewed Papers}
\begin{tabular}{|p{0.7cm}|p{1.2cm}|p{1.7cm}|p{3cm}|p{1cm}|p{1cm}|p{0.7cm}|p{1.5cm}|p{2.1cm}|p{1cm}|}
\hline
\textbf{Paper} & \textbf{Datasets} & \textbf{Preprocessing} & \textbf{Spike Encoding} & \textbf{Neuron Model} & \textbf{\ac{SNN} Arch.} & \textbf{Layers} & \textbf{Training} & \textbf{Hardware \newline Implementation} & \textbf{\ac{ANN} Comparison} \\ \hline
\snn{1} & \ac{CWRU}, \ac{PUD} & - & Current Injection & \ac{LIF} & Feed Forward & 4+ & \ac{ANN} Conversion & No & Yes \\ \hline
\snn{2} & \ac{IBF}, \ac{rtf} & \acp{GFB} & Async delta modulator & \ac{LIF} & Feed Forward & 3 & Unsupervised & DYNAP-SE chip & No \\ \hline
\snn{3} & Custom & \ac{DFT}  & Current Injection & \ac{ALIF} & \acp{LSNN} & 3 & Supervised & No & Yes \\ \hline
\snn{4} & Custom & \ac{FFT}  & Current Injection, \ac{TTFS} & \ac{ALIF} & \acp{LSNN} & 3 & Supervised & STMicroelectronics STM32F407VG6 MCU & No \\ \hline
\snn{5} & \ac{CWRU}, MFPT & \ac{LMD}  & Population Coding, \ac{TTFS} & \ac{LIF} & Feed Forward & 2 & Supervised & No & No \\ \hline
\snn{6} & \ac{CWRU}, Proprietary & \ac{LMD} & Population Coding & \ac{LIF} & Feed Forward & 3 & \ac{ANN} Conversion & No & Yes \\ \hline
\snn{7} & Custom  & \acp{GFB}, \ac{PCA} & Population Coding, \ac{TTFS} & IF & Feed Forward & 3 & Supervised & No & Yes \\ \hline
\snn{8} & UCR & - & Population Coding, \ac{TTFS}, Poisson Rate Coding & \ac{LIF} & Reservoir \ac{SNN} & 2 & Supervised & No & Yes \\ \hline
\snn{9} & \ac{CWRU}, \ac{PUD}, MFPT & \ac{LMD} & Population Coding, Boltzmann & \ac{SRM} & Feed Forward & 4+ & Supervised & No & Yes \\ \hline
\snn{10} & Custom & \ac{SLT} & - & \ac{SRM} & Feed Forward & 2 & Supervised & No & No \\ \hline
\snn{11} & Custom & \acp{GFB}  & Current Injection & \ac{LIF} & Feed Forward & 2 & Unsupervised & No & No\\ \hline
\snn{12} & Proprietary, \ac{CWRU}, Kaggle & \ac{STFT} & - & \ac{adex} & Reservoir \ac{SNN} & 3 & Supervised & No & Yes \\ \hline
\end{tabular}
\label{tab:snn_features}
\end{table*}

\subsection{Discussion}

This subsection will identify and discuss the common features among the listed approaches for \ac{SNN} based \ac{PM}.

\subsubsection{Datasets}
The most commonly used dataset for benchmarking \ac{SNN} performance for \ac{PM} is the \ac{CWRU} dataset \cite{loparo2012case}, with five of the twelve approaches using it (\snn{1}, \snn{5}, \snn{6}, \snn{9}, \snn{12}). Another common approach to evaluating \ac{SNN} performance, used in five papers (\snn{3}, \snn{4}, \snn{7}, \snn{10}, \snn{11}), involves creating custom datasets by recording vibration / acoustic data from various sources, like bridges, viaducts, machinery diagnostic systems and simulations. Less common datasets include the \ac{PUD} \cite{noauthor_konstruktions_nodate} (used by \snn{1} and \snn{9}), the MFPT dataset \cite{noauthor_fault_nodate} (used by \snn{5} and \snn{9}), the \ac{IBF} dataset \cite{bechhoefer2013condition} (used by \snn{2}), the \ac{rtf} dataset \cite{qiu2006wavelet} (used by \snn{2}), real-time classification datasets from the UCR repository \cite{dau2019ucr} (used by \snn{8}) and Kaggle Gearbox Fault Diagnosis dataset \cite{noauthor_gearbox_nodate} (uased by \snn{12}). Proprietary datasets like the deep space detector bearing dataset in the NASA database (used by \snn{6}) and the Petersburg Electrotechnical University dataset (ETU) (used by \snn{12}) are also used.

\subsubsection{Data Preprocessing}
To prepare the raw vibration / acoustic data, preprocessing methods are employed. Most of the approaches employ a time-to-frequency domain conversion. \acfp{GFB} \cite{patterson_efficient_1987} are used in three of the papers (\snn{2}, \snn{7}, \snn{11}). Fourier Transformations are also used in three papers, but in different forms: \ac{DFT} is used in \snn{3}, \ac{FFT} in \snn{4}, and \ac{STFT} is used in \snn{12}. \ac{LMD} is used by the approaches \snn{5}, \snn{6}, \snn{9}. \ac{PCA} is used in \snn{7} and \ac{SLT} is used in \snn{10}.

\subsubsection{Spike Encoding}
Among the presented papers, five of them (\snn{5}, \snn{6}, \snn{7}, \snn{8}, \snn{9}) employ Population Coding with \acfp{GRF} \cite{yu2014brain}. \acf{TTFS} coding is used by four approaches (\snn{4}, \snn{5}, \snn{7}, \snn{8}), as well as Current Injection (\snn{1}, \snn{3}, \snn{4}, \snn{11}). Other used methods include the asynchronous delta modulator \cite{corradi2015neuromorphic} \cite{sharifshazileh2021electronic} (used in \snn{2}), the Poisson Rate Coding (used in \snn{8}) and the Boltzmann distribution based pulse probability sequence conversion (used in \snn{9}). 

\subsubsection{Neuron Models}
The most widely used neuron model is the \ac{LIF} (used in \snn{1}, \snn{2}, \snn{5}, \snn{6}, \snn{8}, \snn{11}). The \ac{ALIF} model is employed in two papers (\snn{3}, \snn{4}), as well as the \ac{SRM} model (\snn{9}, \snn{10}). The \ac{IF} and \ac{adex} models are each used in one paper only (\snn{7} and \snn{12} respectively).

\subsubsection{SNN Architecture}
The most commonly used \ac{SNN} architecture is the feed forward architecture, with eight of the investigated methods using it (\snn{1}, \snn{2}, \snn{5}, \snn{6}, \snn{7}, \snn{9}, \snn{10}, \snn{11}). Two of the papers employ \acp{LSNN} (\snn{3}, \snn{4}) and two others use a Reservoir \ac{SNN} approach (\snn{8}, \snn{12}).

\subsubsection{Model Depth}
Most of the models employ a shallow architecture, by either using only an input and an output layer (\snn{5}, \snn{8}, \snn{10}, \snn{11}), or by using an input, an output and one single hidden layer (\snn{2}, \snn{3}, \snn{4}, \snn{6}, \snn{7}, \snn{12}). Only two papers employ an architecture with two or more hidden layers (\snn{1}, \snn{9}).

\subsubsection{Training}
Most of the listed methods (eight out of twelve) employed a supervised approach to training the \ac{SNN} (\snn{3}, \snn{4}, \snn{5}, \snn{7}, \snn{8}, \snn{9}, \snn{10}, \snn{12}), by use of some of the following approaches: \ac{BPTT} / E-Prop \cite{bellec2020solution}, improved tempotron learning rule \cite{gutig2006tempotron}, margin maximization \cite{dora2021spiking} or logistic regression applied on neuronal traces. Two of the papers (\snn{1}, \snn{6}) applied the conversion technique, by training a \ac{ANN} with a supervised method, then converting it to an equivalent \ac{SNN}. The remaining two papers (\snn{2}, \snn{11}) used unsupervised learning to train their networks.

\subsubsection{Hardware Implementation}
From all of the listed papers, only two of them presented a hardware implementation of the proposed \ac{SNN} (Approach \snn{2}: DYNAP-SE chip \cite{moradi2017scalable}, Approach \snn{4}: STMicroelectronics STM32F407VG6 MCU system-on-chip \cite{noauthor_stm32f407vg_nodate}).

\subsubsection{\ac{ANN} comparison}
Seven papers present their results by comparing them with \acp{ANN} for the same task (\snn{1}, \snn{3}, \snn{6}, \snn{7}, \snn{8}, \snn{9}, \snn{12}).

\subsection{Patents}
Reviewing existing patent applications reveals a need for more descriptions of deploying \acp{SNN} for anomaly detection or \acf{PM} in industrial environments. 

In \cite{US2022026879A1}, the authors describe a system that generates and analyzes a sensor data stream to detect anomalies during a machine's operation. The system comprises one or more sensors, a computation unit with a memory device, and a communication interface. In order to detect anomalies during operation, an \ac{ANN} or an \ac{SNN} is trained with sensor data from a new machine under regular operating conditions. Subsequently, the system is employed to identify instances of behavior that diverge from the training operation patterns.

In \cite{CN115238915A}, \acp{SNN} are integrated into a comprehensive industrial equipment fault prediction and health monitoring system. The system encompasses several submodules, including anomaly detection, fault analysis, and correction mechanisms. The anomaly detection and fault analysis are based on a broad collection of sensor information, including temperature, pressure, noise, vibration, strain, crack, wear, and corrosion. Furthermore, the approach fuses image identification and nondestructive testing to analyze crack, wear, and corrosion damages.

\section{Artificial Neural Networks}
\label{sec:ANN}
\subsection{Predictive Maintenance with Artificial Neural Networks}
Research in \ac{ANN}-based \ac{PM} can be categorized by system architecture, purpose, and approach \cite{RanASurveyofPredictiveMaintenanceSyst2019}. The primary system architectures include Open System Architecture for Condition-based Monitoring, cloud-enhanced \ac{PM}, and \ac{PM} 4.0. \ac{PM} 4.0 provides support for technicians through online analysis of collected data. The Open System Architecture for Condition-based Monitoring, as defined in ISO 13374, offers a standardized, layered framework for \ac{PM} design and implementation. Cloud-enhanced \ac{PM} leverages the potential of cloud computing with a centralized architecture.

From a methodological standpoint, approaches can be classified into three main categories: knowledge-based, traditional \ac{ML}, and \ac{DL}-based techniques. Knowledge-based methods rely on expert knowledge and experience with system faults. Traditional \ac{ML} and \ac{DL} differ primarily in the number of hidden layers within the \ac{NN}. While \ac{ML} uses big data to learn highly nonlinear functions and generalize from similar situations, shallow \acp{ANN} with fewer units and hidden layers struggle to extract hidden information from raw data and require manual feature engineering \cite{RanASurveyofPredictiveMaintenanceSyst2019}. Consequently, \ac{DL} methodologies are typically favored.

The most prominent \ac{DL} methods are \ac{AE}, \ac{CNN}, and \ac{DBN} whereby recurrent architectures are used to keep a memory of the past \cite{RanASurveyofPredictiveMaintenanceSyst2019, HoangAsurveyonDeepLearningbasedbearing2019}. 
Instead of directly applying raw sensor data as input for the \ac{DL} approach, the sensor measurements are preprocessed and transformed into time, frequency, or time-frequency domain to extract features. 

Recent reviews focus on the general application of \ac{DL} methods in \ac{PM} with a focus on architecture, structure, and purpose \cite{RanASurveyofPredictiveMaintenanceSyst2019}. 
A comprehensive survey on \ac{PM} in industry $4.0$ is provided in \cite{CnarMachineLearninginPredictiveMaintenan2020} and with a particular focus on cloud or fog computing in \cite{AngelopoulosTacklingFaultsintheIndustry4.0Era2019}. 
The application of \ac{PM} for bearing fault detecting is considered in \cite{HoangAsurveyonDeepLearningbasedbearing2019, SchwendemannAsurveyofmachinelearningtechniques2021}.
The authors provide an overview of the advances regarding bearing fault diagnosis with \ac{DL} methods.
In \cite{Drakaki2022}, the authors present a survey about fault detection and diagnosis for induction motors.
\ac{DL} methods for engine failure prediction are reviewed in \cite{NamuduriReviewDeepLearningMethodsforSensor2020}.

\subsection{Approaches Emphasizing Low Power}
In \cite{Vitolo2021}, the authors apply a two-stage low-power and in-sensor anomaly detection. 
The architecture is divided into a hardware \ac{AE} and a software \ac{CNN} part.
The \ac{AE} is always on at the sensor and detects anomalies. 
Once an anomaly is detected, the \ac{CNN} is activated and serves as a classifier based on the encoder part of the \ac{AE}.
The results reveal that the system achieves high accuracy with an anomaly detection rate of $99.61 \%$ on the \ac{CWRU} dataset and very low power consumption of the \ac{AE} when implemented on a Xilinx Artix 7 FPGA with $122$mW while operating at the maximum frequency of $45$MHz.

A sensor-fusion \ac{PM} utilizing vibration and sound data for a brushless direct current motor is proposed in \cite{SuawaModelingandFaultDetectionofBrushles2022}.
The \ac{PM} system is deployed on an FPGA in order to allow close hardware-software collaboration and hardware-accelerated algorithms for fault detection. 
The authors investigated the usability of a \ac{CNN}, an \ac{LSTM}, and a combination of both for \ac{PM} besides the comparison of vibration and sound data.
The results generally show that sound information outperforms the vibration in a single-sensor setup, but the accuracy improves with data-level sensor fusion. 
The \ac{CNN} model significantly outperforms the \ac{LSTM} model. 
Combining both models is better than single \ac{LSTM} but less accurate than \ac{CNN}.
The results indicate that the \ac{LSTM} has to be used with a feature extraction step in advance \cite{SuawaModelingandFaultDetectionofBrushles2022}.

A smart vibration sensor for industrial applications with integrated ultra-low power embedded \ac{PM} is proposed in \cite{SebastianMarzettiUltralowPowerEmbeddedUnsupervisedLe2022}.
An unsupervised K-means algorithm based on feature extraction is used for monitoring and failure detection. 
The algorithm is embedded on an ARM M4F with an average power consumption of $80 \mu W$ and leads to one year of battery life with a single CR2032 battery cell.

A decentralized on-device \ac{ML} approach to identify patterns in sensor data is demonstrated in \cite{RenThesynergyofcomplexeventprocessing2021}. 
The approach comprises a framework for distributed sensor networks to shift the computation from the cloud to the edge devices. 
A \ac{CNN} Bi-Directional \ac{LSTM} model for fault prognosis in machines of industry $4.0$ is proposed in \cite{JustusMachinelearningbasedfaultorientedpr2022}.
The method allows the analysis of machine characteristics based on embedded sensors. 
The model is evaluated on the Machine Investigation and Inspection (MIMII) dataset \cite{purohit2019mimii} with reliability over $94\%$.

In \cite{HassanDEEPWINDAnAccurateWindTurbineCondi2020}, the end-to-end multichannel \ac{CNN} condition monitoring and fault detection framework DeepWind for wind turbines is presented.
The DeepWind framework exploits multi-channel \ac{CNN} and can be implemented on resource-constrained embedded devices. 
It detects faults in rotor blades in wind turbines based on automatic feature detection and subsequent classification. 
First, the raw sensor data is downsampled and windowed in a software preprocessing step.
Afterward, the data is transformed into the frequency domain and fed into an embedded Mulit-Channel \ac{CNN}.
The applied \acp{CNN} have $50$ and $40$ filters with kernel sizes of $8$ and $4$.
The system is evaluated on a real wind turbine dataset with a fault detection average of $94\%$.

A study on developing, testing, and evaluating \ac{ML} approaches for low-cost microcontrollers is presented in \cite{GrethlerEmbeddedMachineLearningforMachineCo2021}.
The authors analyze the current state of the art regarding low-power \ac{PM} applications on the edge.
The focus is on algorithms that can be trained and run on limited memory resource devices with online model parameter adaption.
The underlying goal is to avoid needing a separate backend and additional communication.

A self-contained low-power on-device \ac{PM} (LOPdM) system based on a self-powered sensor is elaborated in \cite{ChenLOPdMALowPowerOnDevicePredictive2023}. 
Compared to traditional \ac{PM} systems where the data is transmitted to and processed in a server, the data is locally inferred with low power consumption in the TinyML-based \ac{PM} system.
A dataset with a self-powered sensor from a simulated vibration environment is collected and used for model evaluation.
In the evaluation, random forest and \ac{DNN} model showed the highest accuracy.

In \cite{ChenEcitonVeryLowPowerLSTMNeuralNetwo2021}, the authors present Eciton, a low-power \ac{LSTM} accelerator for \ac{PM} systems with low-power edge-sensor nodes. 
Eciton shows the power consumption of $17$mW under load and reduces memory and chip resources utilizing $8$-bit quantization and sigmoid activation function. 
The accelerator fits on a low-power Lattice iCE40 UP5K FPGA and demonstrates real-time processing with minimal loss of accuracy.
The proposed method is evaluated on two publicly available datasets, a turbofan engine maintenance dataset from NASA \cite{Abhinav2008} and an electrical motor maintenance dataset with vibration and humidity data \cite{Jinyeong2016}. 

A workflow for training a quantized \ac{DL} anomaly detection for devices with limited memory, compute, and power resources is described in \cite{BoonsLowpoweronlinemachinemonitoringat2021}.
A Deep Support Vector Data Description (SVDD) model is used for edge computing to overcome the drawback of large \ac{AE} composed of an encoder and decoder, resulting in many layers. 
The detection performance of the SVDD model is evaluated in terms of AUC on the MIMII dataset.
The proposed SVDD model outperforms traditional \acp{AE} and reduces the computational complexity by $50\%$.

An online low-power signal processing algorithm for fault detection based on frequency data is shown in \cite{AsadACurrentSpectrumBasedAlgorithmforF2023}.
The algorithm can improve the detection of small amplitudes at fault representing frequencies without complex signal processing and is therefore suitable for on-device applications.

\subsection{Discussion}
A comparison of different approaches with regard to characteristic features is shown in Table \ref{tab:ann_features}.
Most approaches use custom data recorded with individual sensors to evaluate their proposed approach. 
Two approaches use the MIMII dataset, one approach the \ac{CWRU}, and one approach a turbofan engine dataset from NASA as well as an electrical motor dataset. 
A common pre-processing technique across the reviewed paper is to transform the time series data into the frequency domain using Fourier Transformation. 
For this, either a \ac{FFT} or a \ac{STFT} is used.
Additionally, the input data can be windowed to process the data in patterns. 
This can also be used to obtain a fixed input size for the \ac{ANN}. 
As network architecture, \ac{AE} and recurrent approaches are often used.
All recurrent approaches use an \ac{LSTM} to process time series data and keep track of the history. 
Convolutional operations are used to extract features. 
Either as part of the \ac{AE} or as an additional feature extractor. 
$7$ out of $8$ approaches are trained in a supervised manner and one in an unsupervised. 
None of the approaches compares the performance of the \ac{ANN} with an \ac{SNN} approach.

\begin{table*}
\centering
\caption{ANN Features in Reviewed Papers}
\begin{tabular}{|p{0.7cm}|p{1cm}|p{2.6cm}|p{1.3cm}|p{1.4cm}|p{1.5cm}|p{0.9cm}|p{1.5cm}|p{2.5cm}|}
\hline
\textbf{Paper} & \textbf{Datasets} & \textbf{Preprocessing} & \textbf{Activation} & \textbf{Architecture} & \textbf{ANN Model} & \textbf{Layers} & \textbf{Training} & \textbf{Hardware \newline Implementation}  \\ \hline
\cite{Vitolo2021} &  \ac{CWRU} &  No &   ReLU    &  Classifier, \ac{AE}  & \ac{CNN}   &  6 &  Supervised  &  Xilinx Artix 7 FPGA    \\ \hline
\cite{SuawaModelingandFaultDetectionofBrushles2022} & Custom & Raw data segmentation & ReLU &  Classifier, Recurrent &\ac{CNN}, \ac{CNN}-\ac{LSTM}, \ac{LSTM}& 5+ &  Supervised & FPGA \\ \hline
\cite{SebastianMarzettiUltralowPowerEmbeddedUnsupervisedLe2022} & Custom & \ac{FFT} & - &  Classifier & -  &  -  & Unsupervised &  TI CC2652  \\ \hline
\cite{JustusMachinelearningbasedfaultorientedpr2022}& MIMII & LPC & ReLU  & Classifier, Recurrent & \ac{CNN}-\ac{LSTM} & 8 & Supervised & -      \\ \hline
\cite{HassanDEEPWINDAnAccurateWindTurbineCondi2020}& Custom & Windowing, \ac{FFT} & - &  Classifier  &  \ac{CNN}, Feed Forward  & 5 & Supervised & FPGA   \\ \hline
\cite{ChenLOPdMALowPowerOnDevicePredictive2023}&  Custom & \ac{FFT} & ReLU & Classifier & Feed Forwad  &  1-4 & Supervised &  ESP32 -Tensilica Xtensa LX6   \\ \hline
\cite{ChenEcitonVeryLowPowerLSTMNeuralNetwo2021}& NASA \cite{Abhinav2008}, Electric motor \cite{Jinyeong2016} &  -  &  Sigmoid, Tanh &  Accelerator, Recurrent  & \ac{LSTM} &  3-4  & Supervised & Lattice iCE40 UP5K FPGA   \\ \hline
\cite{BoonsLowpoweronlinemachinemonitoringat2021}& MIMII & \ac{STFT} & ReLU, Linear  & Classifier, \ac{AE}  &  Feed Forward, SVDD   & 6 & Supervised & -    \\ \hline
\end{tabular}
\label{tab:ann_features}
\end{table*}

\subsection{Patents}
The survey is extended to patents publications in the field of ANNs with an emphasis on low-power predictive maintenance applications. Two patent specifications are published that disclose applications in the given context.

In \cite{CN115941729A}, the authors propose the TinyML technology \cite{RAY20221595} for use in predictive maintenance applications.
The goal of the development is the shift of data processing from the cloud to the edge to enhance data efficiency.
With a single-chip microcomputer as the computation platform, the authors achieve a low-cost and low-power \ac{ANN} deployment to analyze industrial equipment on the edge.

In \cite{CN116184900A}, the patent claim in \cite{CN115941729A} is extended with kinetic energy harvesting to promote a durable application without the need for external energy.
In this regard, the authors deploy a piezoelectric sensor as the information source and the main energy supply.
Consequently, the invention allows for the application in extreme environments with limited energy resources, in which changing the battery causes additional efforts.

\section{ANN-SNN Comparison}
A comparison between the two fundamental approaches can be made based on the papers analyzed in this work concerning both \acp{SNN} \ref{sec:SNN} and \acp{ANN} \ref{sec:ANN}.
The most common method to evaluate performance is to create custom datasets. The most used publicly available dataset for both approaches is the \ac{CWRU} dataset \cite{loparo2012case}.
The most widely employed preprocessing method for \acp{SNN} and \acp{ANN} is a transformation from the time domain to the frequency domain.
The most used \ac{SNN} architecture is a simple Feed Forward, with few approaches employing more complex structures such as \acp{LSNN} or Reservoirs. On the other hand, \acp{ANN} employ various strategies such as \ac{CNN}, \ac{LSTM}, and Feed Forward equally. One possible reason may be the inherent recurrent nature of \acp{SNN} resulting from the stateful neuron units.
\acp{SNN} more often employ shallow architectures (1-3 layers), whereas \acp{ANN} focus on deeper models (4+). This could be due to the challenges associated with training of deep \acp{SNN}.
The vast majority of all the approaches present a supervised training method.

\section{Conclusion}
The paper's main objective is to explore the use of \acfp{NN} for low-power \acf{PM} in Industry 4.0. The motivation for this work stems from some of the drawbacks of the traditional approaches to \ac{PM}, which include the high transmission and storage costs of analyzing sensor data in the cloud.

This work reviews the existing literature on \acfp{SNN} for \ac{PM} concerning training datasets, preprocessing, spike encoding, neuron models, \ac{SNN} architecture, layer depth, training methods, and hardware implementations. The analysis extends to the most prominent \acf{ANN} approaches regarding datasets, preprocessing, network architecture and models, depth, training methods, and hardware implementation. A comparison between \acp{ANN} and \acp{SNN} finds that both approaches often use the same preprocessing methods, primarily focused on transformations from the time domain to the frequency domain. \ac{ANN} models and architectures are more diverse, exploring a mix of \acp{CNN}, \ac{LSTM}, and Feed-Forward models with deeper networks. In contrast, \ac{SNN} approaches mainly employ Feed-Forward architectures and shallow networks. The comparison also finds no agreement towards a standard benchmark dataset for predictive maintenance within or across the two fundamental methods. Limited information is available regarding hardware implementations for both methods.

This paper suggests that future research in low-power \ac{PM} should focus on practical hardware implementations of neural networks. Furthermore, since most of the analyzed methods recoursed to creating custom datasets, it suggests that a standard benchmark for low-power predictive maintenance is needed.

\section*{Disclosure}
We employed grammar-checking software to identify and correct grammatical errors, typos, and stylistic inconsistencies (Grammarly \cite{noauthor_grammarly_nodate}, DeepL Write \cite{noauthor_deepl_nodate}).

\section*{Acknowledgments}
This research is funded by the German Federal Ministry of Education and Research as part of the project ”ThinKIsense“, funding no. 16ME0564.

\bibliographystyle{IEEEtran}
\bibliography{literature}

\begin{thebibliography}{10}
\providecommand{\url}[1]{#1}
\csname url@samestyle\endcsname
\providecommand{\newblock}{\relax}
\providecommand{\bibinfo}[2]{#2}
\providecommand{\BIBentrySTDinterwordspacing}{\spaceskip=0pt\relax}
\providecommand{\BIBentryALTinterwordstretchfactor}{4}
\providecommand{\BIBentryALTinterwordspacing}{\spaceskip=\fontdimen2\font plus
\BIBentryALTinterwordstretchfactor\fontdimen3\font minus \fontdimen4\font\relax}
\providecommand{\BIBforeignlanguage}[2]{{%
\expandafter\ifx\csname l@#1\endcsname\relax
\typeout{** WARNING: IEEEtran.bst: No hyphenation pattern has been}%
\typeout{** loaded for the language `#1'. Using the pattern for}%
\typeout{** the default language instead.}%
\else
\language=\csname l@#1\endcsname
\fi
#2}}
\providecommand{\BIBdecl}{\relax}
\BIBdecl

\bibitem{noauthor_smart_nodate}
\BIBentryALTinterwordspacing
``\BIBforeignlanguage{en}{Smart {Sensors} {Market} - {Forecast}, {Share} \& {Growth}}.'' [Online]. Available: \url{https://www.mordorintelligence.com/industry-reports/global-smart-sensors-market-industry}
\BIBentrySTDinterwordspacing

\bibitem{vishwakarma2017vibration}
M.~Vishwakarma, R.~Purohit, V.~Harshlata, and P.~Rajput, ``Vibration analysis \& condition monitoring for rotating machines: a review,'' \emph{Materials Today: Proceedings}, vol.~4, no.~2, pp. 2659--2664, 2017.

\bibitem{jombo2023acoustic}
G.~Jombo and Y.~Zhang, ``Acoustic-based machine condition monitoring—methods and challenges,'' \emph{Eng}, vol.~4, no.~1, pp. 47--79, 2023.

\bibitem{1noauthor_stmicroelectronics_nodate}
\BIBentryALTinterwordspacing
``{STMicroelectronics}.'' [Online]. Available: \url{https://diolan.com/st}
\BIBentrySTDinterwordspacing

\bibitem{2noauthor_efr32bg27_nodate}
\BIBentryALTinterwordspacing
``\BIBforeignlanguage{en}{{EFR32BG27} {Series} 2 {Small} {Bluetooth} {LE} {SoC} - {Silicon} {Labs}}.'' [Online]. Available: \url{https://www.silabs.com/wireless/bluetooth/efr32bg27-series-2-socs}
\BIBentrySTDinterwordspacing

\bibitem{3noauthor_tidc-cc2650stk-sensortag_nodate}
\BIBentryALTinterwordspacing
``{TIDC}-{CC2650STK}-{SENSORTAG} reference design {TI}.com.'' [Online]. Available: \url{https://www.ti.com/tool/TIDC-CC2650STK-SENSORTAG}
\BIBentrySTDinterwordspacing

\bibitem{4noauthor_lsm6dsl_nodate}
\BIBentryALTinterwordspacing
``{LSM6DSL} - {iNEMO} {6DoF} inertial measurement unit ({IMU}), for smart phones and battery operated {IoT}, {Gaming}, {Wearable} and {Consumer} {Electronics}. {Ultra}-low power and high accuracy - {STMicroelectronics}.'' [Online]. Available: \url{https://www.st.com/en/mems-and-sensors/lsm6dsl.html}
\BIBentrySTDinterwordspacing

\bibitem{5noauthor_new_nodate}
\BIBentryALTinterwordspacing
``Knowles {Ultrasonic} {MEMS} {Microphone}.'' [Online]. Available: \url{https://www.knowles.com/resource-news-details/new-product-ultrasonic-mems-microphone}
\BIBentrySTDinterwordspacing

\bibitem{6noauthor_sitrans_nodate}
\BIBentryALTinterwordspacing
``\BIBforeignlanguage{en}{{SITRANS} {SCM} {IQ}}.'' [Online]. Available: \url{https://www.siemens.com/global/en/products/automation/process-instrumentation/digitalization/smart-condition-monitoring.html}
\BIBentrySTDinterwordspacing

\bibitem{7noauthor_condition_nodate}
\BIBentryALTinterwordspacing
``Condition {Monitoring} for rotating equipment - {Motor} condition monitoring{\textbar} {ABB} {Motion} {Services} - {ABB} {Motion} {Services} {\textbar} {Data} and {Advisory} services ({ABB} {Motion} {Services}).'' [Online]. Available: \url{https://new.abb.com/service/motion/data-and-advisory-services/condition-monitoring-for-rotating-equipment}
\BIBentrySTDinterwordspacing

\bibitem{8noauthor_honeywell_nodate}
\BIBentryALTinterwordspacing
``\BIBforeignlanguage{en-US}{Honeywell {Versatilis}™ {Transmitter} {\textbar} {Honeywell}}.'' [Online]. Available: \url{https://process.honeywell.com/us/en/products/field-instruments/honeywell-versatilis-equipment-health-monitoring-ehm-solutions/honeywell-versatilis-transmitter}
\BIBentrySTDinterwordspacing

\bibitem{9corporation_xs770a_2019}
\BIBentryALTinterwordspacing
Y.~E. Corporation, ``\BIBforeignlanguage{en}{{XS770A} {Wireless} {Vibration} {Sensor}},'' 2019. [Online]. Available: \url{https://web-material3.yokogawa.com/13/23403/file_name/GS01W06E01-01EN.pdf}
\BIBentrySTDinterwordspacing

\bibitem{li_ebc_pm_2024}
X.~Li, S.~Yu, Y.~Lei, N.~Li, and B.~Yang, ``Intelligent machinery fault diagnosis with event-based camera,'' \emph{IEEE Transactions on Industrial Informatics}, vol.~20, no.~1, pp. 380--389, 2024.

\bibitem{lenk2023neuromorphic}
C.~Lenk, P.~H{\"o}vel, K.~Ved, S.~Durstewitz, T.~Meurer, T.~Fritsch, A.~M{\"a}nnchen, J.~K{\"u}ller, D.~Beer, T.~Ivanov \emph{et~al.}, ``Neuromorphic acoustic sensing using an adaptive microelectromechanical cochlea with integrated feedback,'' \emph{Nature Electronics}, vol.~6, no.~5, pp. 370--380, 2023.

\bibitem{auge2021survey}
D.~Auge, J.~Hille, E.~Mueller, and A.~Knoll, ``A survey of encoding techniques for signal processing in spiking neural networks,'' \emph{Neural Processing Letters}, vol.~53, no.~6, pp. 4693--4710, 2021.

\bibitem{schuman2019non}
C.~D. Schuman, J.~S. Plank, G.~Bruer, and J.~Anantharaj, ``Non-traditional input encoding schemes for spiking neuromorphic systems,'' in \emph{2019 International Joint Conference on Neural Networks (IJCNN)}.\hskip 1em plus 0.5em minus 0.4em\relax IEEE, 2019, pp. 1--10.

\bibitem{schuman2022evaluating}
C.~Schuman, C.~Rizzo, J.~McDonald-Carmack, N.~Skuda, and J.~Plank, ``Evaluating encoding and decoding approaches for spiking neuromorphic systems,'' in \emph{Proceedings of the International Conference on Neuromorphic Systems 2022}, 2022, pp. 1--9.

\bibitem{nitzsche_ultra-low_2021}
S.~Nitzsche, M.~Neher, S.~von Dosky, and J.~Becker, ``Ultra-low power machinery fault detection using deep neural networks,'' in \emph{Machine Learning and Principles and Practice of Knowledge Discovery in Databases}, ser. Communications in Computer and Information Science.\hskip 1em plus 0.5em minus 0.4em\relax Springer International Publishing, 2021, pp. 390--396.

\bibitem{loparo2012case}
K.~Loparo, ``Case western reserve university bearing data center,'' \emph{Bearings Vibration Data Sets, Case Western Reserve University}, pp. 22--28, 2012.

\bibitem{noauthor_konstruktions_nodate}
\BIBentryALTinterwordspacing
Konstruktions- und antriebstechnik ({KAt}) - bearing {DataCenter}(universität paderborn). [Online]. Available: \url{https://mb.uni-paderborn.de/kat/forschung/kat-datacenter/bearing-datacenter}
\BIBentrySTDinterwordspacing

\bibitem{bekolay2014nengo}
T.~Bekolay, J.~Bergstra, E.~Hunsberger, T.~DeWolf, T.~C. Stewart, D.~Rasmussen, X.~Choo, A.~R. Voelker, and C.~Eliasmith, ``Nengo: a python tool for building large-scale functional brain models,'' \emph{Frontiers in neuroinformatics}, vol.~7, p.~48, 2014.

\bibitem{dennler_online_2021}
\BIBentryALTinterwordspacing
N.~Dennler, G.~Haessig, M.~Cartiglia, and G.~Indiveri, ``Online detection of vibration anomalies using balanced spiking neural networks,'' in \emph{2021 {IEEE} 3rd International Conference on Artificial Intelligence Circuits and Systems ({AICAS})}.\hskip 1em plus 0.5em minus 0.4em\relax {IEEE}, 2021, pp. 1--4. [Online]. Available: \url{https://ieeexplore.ieee.org/document/9458403/}
\BIBentrySTDinterwordspacing

\bibitem{bourdoukan2012learning}
R.~Bourdoukan, D.~Barrett, S.~Deneve, and C.~K. Machens, ``Learning optimal spike-based representations,'' \emph{Advances in neural information processing systems}, vol.~25, 2012.

\bibitem{stimberg2019brian}
M.~Stimberg, R.~Brette, and D.~F. Goodman, ``Brian 2, an intuitive and efficient neural simulator,'' \emph{elife}, vol.~8, p. e47314, 2019.

\bibitem{moradi2017scalable}
S.~Moradi, N.~Qiao, F.~Stefanini, and G.~Indiveri, ``A scalable multicore architecture with heterogeneous memory structures for dynamic neuromorphic asynchronous processors (dynaps),'' \emph{IEEE transactions on biomedical circuits and systems}, vol.~12, no.~1, pp. 106--122, 2017.

\bibitem{bechhoefer2013condition}
E.~Bechhoefer, ``Condition based maintenance fault database for testing diagnostics and prognostic algorithms,'' \emph{MFPT Data}, 2013.

\bibitem{qiu2006wavelet}
H.~Qiu, J.~Lee, J.~Lin, and G.~Yu, ``Wavelet filter-based weak signature detection method and its application on rolling element bearing prognostics,'' \emph{Journal of sound and vibration}, vol. 289, no. 4-5, pp. 1066--1090, 2006.

\bibitem{patterson_efficient_1987}
\BIBentryALTinterwordspacing
R.~PATTERSON, ``An efficient auditory filterbank based on the gammatone function,'' \emph{Meeting of the IOC Speech Group on Auditory Modelling at RSRE, 1987}, 1987. [Online]. Available: \url{https://cir.nii.ac.jp/crid/1570291225803648384}
\BIBentrySTDinterwordspacing

\bibitem{corradi2015neuromorphic}
F.~Corradi and G.~Indiveri, ``A neuromorphic event-based neural recording system for smart brain-machine-interfaces,'' \emph{IEEE transactions on biomedical circuits and systems}, vol.~9, no.~5, pp. 699--709, 2015.

\bibitem{sharifshazileh2021electronic}
M.~Sharifshazileh, K.~Burelo, J.~Sarnthein, and G.~Indiveri, ``An electronic neuromorphic system for real-time detection of high frequency oscillations (hfo) in intracranial eeg,'' \emph{Nature communications}, vol.~12, no.~1, p. 3095, 2021.

\bibitem{zanatta_damage_2021}
\BIBentryALTinterwordspacing
L.~Zanatta, F.~Barchi, A.~Burrello, A.~Bartolini, D.~Brunelli, and A.~Acquaviva, ``Damage detection in structural health monitoring with spiking neural networks,'' in \emph{2021 {IEEE} International Workshop on Metrology for Industry 4.0 \& {IoT} ({MetroInd}4.0\&{IoT})}.\hskip 1em plus 0.5em minus 0.4em\relax {IEEE}, 2021, pp. 105--110. [Online]. Available: \url{https://ieeexplore.ieee.org/document/9488476/}
\BIBentrySTDinterwordspacing

\bibitem{bellec2020solution}
G.~Bellec, F.~Scherr, A.~Subramoney, E.~Hajek, D.~Salaj, R.~Legenstein, and W.~Maass, ``A solution to the learning dilemma for recurrent networks of spiking neurons,'' \emph{Nature communications}, vol.~11, no.~1, p. 3625, 2020.

\bibitem{barchi_spiking_2021}
\BIBentryALTinterwordspacing
F.~Barchi, L.~Zanatta, E.~Parisi, A.~Burrello, D.~Brunelli, A.~Bartolini, and A.~Acquaviva, ``Spiking neural network-based near-sensor computing for damage detection in structural health monitoring,'' \emph{Future Internet}, vol.~13, no.~8, p. 219, 2021. [Online]. Available: \url{https://www.mdpi.com/1999-5903/13/8/219}
\BIBentrySTDinterwordspacing

\bibitem{noauthor_stm32f407vg_nodate}
\BIBentryALTinterwordspacing
``\BIBforeignlanguage{en}{{STM32F407VG} - {High}-performance foundation line, {Arm} {Cortex}-{M4} core with {DSP} and {FPU}, 1 {Mbyte} of {Flash} memory, 168 {MHz} {CPU}, {ART} {Accelerator}, {Ethernet}, {FSMC} - {STMicroelectronics}}.'' [Online]. Available: \url{https://www.st.com/en/microcontrollers-microprocessors/stm32f407vg.html}
\BIBentrySTDinterwordspacing

\bibitem{zuo_spiking_2021}
\BIBentryALTinterwordspacing
L.~Zuo, L.~Zhang, Z.-H. Zhang, X.-L. Luo, and Y.~Liu, ``A spiking neural network-based approach to bearing fault diagnosis,'' \emph{Journal of Manufacturing Systems}, vol.~61, pp. 714--724, 2021. [Online]. Available: \url{https://linkinghub.elsevier.com/retrieve/pii/S0278612520301138}
\BIBentrySTDinterwordspacing

\bibitem{noauthor_fault_nodate}
\BIBentryALTinterwordspacing
``\BIBforeignlanguage{en-US}{Fault {Data} {Sets}}.'' [Online]. Available: \url{https://www.mfpt.org/fault-data-sets/}
\BIBentrySTDinterwordspacing

\bibitem{yu2014brain}
Q.~Yu, H.~Tang, K.~C. Tan, and H.~Yu, ``A brain-inspired spiking neural network model with temporal encoding and learning,'' \emph{Neurocomputing}, vol. 138, pp. 3--13, 2014.

\bibitem{gutig2006tempotron}
R.~G{\"u}tig and H.~Sompolinsky, ``The tempotron: a neuron that learns spike timing--based decisions,'' \emph{Nature neuroscience}, vol.~9, no.~3, pp. 420--428, 2006.

\bibitem{li_research_2022}
\BIBentryALTinterwordspacing
R.~Li and J.~Yuan, ``Research on fault diagnosis based on spiking neural networks in deep space environment,'' in \emph{2022 3rd Asia Service Sciences and Software Engineering Conference}.\hskip 1em plus 0.5em minus 0.4em\relax {ACM}, 2022, pp. 165--170. [Online]. Available: \url{https://dl.acm.org/doi/10.1145/3523181.3523205}
\BIBentrySTDinterwordspacing

\bibitem{zhang_machine_2022}
\BIBentryALTinterwordspacing
Y.~Zhang, S.~Dora, M.~Martinez-Garcia, and S.~Bhattacharyaand, ``Machine hearing for industrial acoustic monitoring using cochleagram and spiking neural network,'' in \emph{2022 {IEEE}/{ASME} International Conference on Advanced Intelligent Mechatronics ({AIM})}.\hskip 1em plus 0.5em minus 0.4em\relax {IEEE}, 2022, pp. 1047--1051. [Online]. Available: \url{https://ieeexplore.ieee.org/document/9863412/}
\BIBentrySTDinterwordspacing

\bibitem{noauthor_products_nodate}
\BIBentryALTinterwordspacing
``Pt 500 machinery diagnostic system.'' [Online]. Available: \url{https://www.gunt.de/index.php?option=com_gunt&task=gunt.list.category&product_id=1022&lang=en&Itemid=148}
\BIBentrySTDinterwordspacing

\bibitem{dora2021spiking}
S.~Dora and N.~Kasabov, ``Spiking neural networks for computational intelligence: an overview,'' \emph{Big Data and Cognitive Computing}, vol.~5, no.~4, p.~67, 2021.

\bibitem{dey_efficient_2022}
\BIBentryALTinterwordspacing
S.~Dey, D.~Banerjee, A.~M. George, A.~Mukherjee, and A.~Pal, ``Efficient time series classification using spiking reservoir,'' in \emph{2022 International Joint Conference on Neural Networks ({IJCNN})}.\hskip 1em plus 0.5em minus 0.4em\relax {IEEE}, 2022, pp. 1--8. [Online]. Available: \url{https://ieeexplore.ieee.org/document/9892728/}
\BIBentrySTDinterwordspacing

\bibitem{dau2019ucr}
H.~A. Dau, A.~Bagnall, K.~Kamgar, C.-C.~M. Yeh, Y.~Zhu, S.~Gharghabi, C.~A. Ratanamahatana, and E.~Keogh, ``The ucr time series archive,'' \emph{IEEE/CAA Journal of Automatica Sinica}, vol.~6, no.~6, pp. 1293--1305, 2019.

\bibitem{pal2020instant}
A.~Pal, A.~Ukil, T.~Deb, I.~Sahu, and A.~Majumdar, ``Instant adaptive learning: An adaptive filter based fast learning model construction for sensor signal time series classification on edge devices,'' in \emph{ICASSP 2020-2020 IEEE International Conference on Acoustics, Speech and Signal Processing (ICASSP)}.\hskip 1em plus 0.5em minus 0.4em\relax IEEE, 2020, pp. 8339--8343.

\bibitem{malhotra2017timenet}
P.~Malhotra, V.~TV, L.~Vig, P.~Agarwal, and G.~Shroff, ``Timenet: Pre-trained deep recurrent neural network for time series classification,'' \emph{arXiv preprint arXiv:1706.08838}, 2017.

\bibitem{zuo_multi-layer_2022}
\BIBentryALTinterwordspacing
L.~Zuo, F.~Xu, C.~Zhang, T.~Xiahou, and Y.~Liu, ``A multi-layer spiking neural network-based approach to bearing fault diagnosis,'' \emph{Reliability Engineering \& System Safety}, vol. 225, p. 108561, 2022. [Online]. Available: \url{https://linkinghub.elsevier.com/retrieve/pii/S0951832022002095}
\BIBentrySTDinterwordspacing

\bibitem{ali_novel_2022}
\BIBentryALTinterwordspacing
Y.~H. Ali, F.~Y. H.~Ahmed, A.~M. Abdelrhman, S.~M. Ali, A.~A. Borhana, and R.~Ishak Raja~Hamzah, ``Novel spiking neural network model for gear fault diagnosis,'' in \emph{2022 2nd International Conference on Emerging Smart Technologies and Applications ({eSmarTA})}.\hskip 1em plus 0.5em minus 0.4em\relax {IEEE}, 2022, pp. 1--6. [Online]. Available: \url{https://ieeexplore.ieee.org/document/9935414/}
\BIBentrySTDinterwordspacing

\bibitem{joseph_spiking_2022}
\BIBentryALTinterwordspacing
G.~V. Joseph and V.~Pakrashi, ``Spiking neural networks for structural health monitoring,'' \emph{Sensors}, vol.~22, no.~23, p. 9245, 2022. [Online]. Available: \url{https://www.mdpi.com/1424-8220/22/23/9245}
\BIBentrySTDinterwordspacing

\bibitem{eliasmith2003neural}
C.~Eliasmith and C.~H. Anderson, \emph{Neural engineering: Computation, representation, and dynamics in neurobiological systems}.\hskip 1em plus 0.5em minus 0.4em\relax MIT press, 2003.

\bibitem{kholkin_comparing_2023}
\BIBentryALTinterwordspacing
V.~Kholkin, O.~Druzhina, V.~Vatnik, M.~Kulagin, T.~Karimov, and D.~Butusov, ``Comparing reservoir artificial and spiking neural networks in machine fault detection tasks,'' \emph{{BDCC}}, vol.~7, no.~2, p. 110, 2023. [Online]. Available: \url{https://www.mdpi.com/2504-2289/7/2/110}
\BIBentrySTDinterwordspacing

\bibitem{noauthor_gearbox_nodate}
\BIBentryALTinterwordspacing
``\BIBforeignlanguage{en}{Gearbox {Fault} {Diagnosis}}.'' [Online]. Available: \url{https://www.kaggle.com/datasets/brjapon/gearbox-fault-diagnosis}
\BIBentrySTDinterwordspacing

\bibitem{gallicchio2013tree}
C.~Gallicchio and A.~Micheli, ``Tree echo state networks,'' \emph{Neurocomputing}, vol. 101, pp. 319--337, 2013.

\bibitem{maass2002real}
W.~Maass, T.~Natschl{\"a}ger, and H.~Markram, ``Real-time computing without stable states: A new framework for neural computation based on perturbations,'' \emph{Neural computation}, vol.~14, no.~11, pp. 2531--2560, 2002.

\bibitem{brette2005adaptive}
R.~Brette and W.~Gerstner, ``Adaptive exponential integrate-and-fire model as an effective description of neuronal activity,'' \emph{Journal of neurophysiology}, vol.~94, no.~5, pp. 3637--3642, 2005.

\bibitem{US2022026879A1}
\BIBentryALTinterwordspacing
``\BIBforeignlanguage{en}{Predictive maintenance of components used in machine automation}.'' [Online]. Available: \url{https://worldwide.espacenet.com/patent/search/family/079586302/publication/US2022026879A1?q=US2022026879A1}
\BIBentrySTDinterwordspacing

\bibitem{CN115238915A}
\BIBentryALTinterwordspacing
``\BIBforeignlanguage{en}{Industrial equipment fault prediction and health monitoring system}.'' [Online]. Available: \url{https://worldwide.espacenet.com/patent/search/family/083670292/publication/CN115238915A?q=pn\%3DCN115238915A}
\BIBentrySTDinterwordspacing

\bibitem{RanASurveyofPredictiveMaintenanceSyst2019}
Y.~Ran, X.~Zhou, P.~Lin, Y.~Wen, and R.~Deng, ``A survey of predictive maintenance: Systems, purposes and approaches,'' 12.12.2019.

\bibitem{HoangAsurveyonDeepLearningbasedbearing2019}
D.-T. Hoang and H.-J. Kang, ``A survey on deep learning based bearing fault diagnosis,'' \emph{Neurocomputing}, vol. 335, pp. 327--335, 2019.

\bibitem{CnarMachineLearninginPredictiveMaintenan2020}
Z.~M. {\c{C}}{\i}nar, A.~{Abdussalam Nuhu}, Q.~Zeeshan, O.~Korhan, M.~Asmael, and B.~Safaei, ``Machine learning in predictive maintenance towards sustainable smart manufacturing in industry 4.0,'' \emph{Sustainability}, vol.~12, no.~19, p. 8211, 2020.

\bibitem{AngelopoulosTacklingFaultsintheIndustry4.0Era2019}
A.~Angelopoulos, E.~T. Michailidis, N.~Nomikos, P.~Trakadas, A.~Hatziefremidis, S.~Voliotis, and T.~Zahariadis, ``Tackling faults in the industry 4.0 era-a survey of machine-learning solutions and key aspects,'' \emph{Sensors (Basel, Switzerland)}, vol.~20, no.~1, 2019.

\bibitem{SchwendemannAsurveyofmachinelearningtechniques2021}
S.~Schwendemann, Z.~Amjad, and A.~Sikora, ``A survey of machine-learning techniques for condition monitoring and predictive maintenance of bearings in grinding machines,'' \emph{Computers in Industry}, vol. 125, p. 103380, 2021.

\bibitem{Drakaki2022}
M.~Drakaki, Y.~L. Karnavas, I.~A. Tziafettas, V.~Linardos, and P.~Tzionas, ``Machine learning and deep learning based methods toward industry 4.0 predictive maintenance in induction motors: State of the art survey,'' \emph{Journal of Industrial Engineering and Management}, vol.~15, no.~1, p.~31, 2022.

\bibitem{NamuduriReviewDeepLearningMethodsforSensor2020}
S.~Namuduri, B.~N. Narayanan, V.~S.~P. Davuluru, L.~Burton, and S.~Bhansali, ``Review---deep learning methods for sensor based predictive maintenance and future perspectives for electrochemical sensors,'' \emph{Journal of The Electrochemical Society}, vol. 167, no.~3, p. 037552, 2020.

\bibitem{Vitolo2021}
P.~Vitolo, G.~D. Licciardo, L.~{Di Benedetto}, R.~Liguori, A.~Rubino, and D.~Pau, ``Low-power anomaly detection and classification system based on a partially binarized autoencoder for in-sensor computing,'' in \emph{2021 28th IEEE International Conference on Electronics, Circuits, and Systems (ICECS)}.\hskip 1em plus 0.5em minus 0.4em\relax IEEE, 2021, pp. 1--5.

\bibitem{SuawaModelingandFaultDetectionofBrushles2022}
P.~Suawa, T.~Meisel, M.~Jongmanns, M.~Huebner, and M.~Reichenbach, ``Modeling and fault detection of brushless direct current motor by deep learning sensor data fusion,'' \emph{Sensors (Basel, Switzerland)}, vol.~22, no.~9, 2022.

\bibitem{SebastianMarzettiUltralowPowerEmbeddedUnsupervisedLe2022}
{Sebasti{\'a}n Marzetti}, {Valentin Gies}, {Valentin Barchasz}, {Herv{\'e} Barth{\'e}lemy}, and {Herv{\'e} Glotin}, ``Ultra-low power embedded unsupervised learning smart sensor for industrial fault detection,'' \emph{2020 IEEE International Conference on Internet of Things and Intelligence System (IoTaIS)}, 2022.

\bibitem{RenThesynergyofcomplexeventprocessing2021}
H.~Ren, D.~Anicic, and T.~A. Runkler, ``The synergy of complex event processing and tiny machine learning in industrial iot,'' in \emph{Proceedings of the 15th ACM International Conference on Distributed and Event-based Systems}, A.~Margara and E.~{Della Valle}, Eds.\hskip 1em plus 0.5em minus 0.4em\relax ACM, 2021, pp. 126--135.

\bibitem{JustusMachinelearningbasedfaultorientedpr2022}
V.~Justus and G.~R. Kanagachidambaresan, ``Machine learning based fault-oriented predictive maintenance in industry 4.0,'' \emph{International Journal of System Assurance Engineering and Management}, vol.~15, no.~1, pp. 462--474, 2022.

\bibitem{purohit2019mimii}
H.~Purohit, R.~Tanabe, K.~Ichige, T.~Endo, Y.~Nikaido, K.~Suefusa, and Y.~Kawaguchi, ``Mimii dataset: Sound dataset for malfunctioning industrial machine investigation and inspection,'' 2019.

\bibitem{HassanDEEPWINDAnAccurateWindTurbineCondi2020}
G.~M. Hassan, A.~Rahil, R.~Sneha, M.~Suraj, K.~Maurice, and P.~J. Felix, \emph{DEEPWIND: An Accurate Wind Turbine Condition Monitoring Framework via Deep Learning on Embedded Platforms}.\hskip 1em plus 0.5em minus 0.4em\relax Piscataway, NJ: IEEE, 2020.

\bibitem{GrethlerEmbeddedMachineLearningforMachineCo2021}
M.~Grethler, M.~B. Marinov, and V.~Klumpp, ``Embedded machine learning for machine condition monitoring,'' in \emph{Future Access Enablers for Ubiquitous and Intelligent Infrastructures}, ser. Lecture Notes of the Institute for Computer Sciences, Social Informatics and Telecommunications Engineering, D.~Perakovic and L.~Knapcikova, Eds.\hskip 1em plus 0.5em minus 0.4em\relax {Springer International Publishing}, 2021, vol. 382, pp. 217--228.

\bibitem{ChenLOPdMALowPowerOnDevicePredictive2023}
Z.~Chen, Y.~Gao, and J.~Liang, ``Lopdm: A low-power on-device predictive maintenance system based on self-powered sensing and tinyml,'' \emph{IEEE Transactions on Instrumentation and Measurement}, vol.~72, pp. 1--13, 2023.

\bibitem{ChenEcitonVeryLowPowerLSTMNeuralNetwo2021}
J.~Chen, S.~Hong, W.~He, J.~Moon, and S.-W. Jun, ``Eciton: Very low-power lstm neural network accelerator for predictive maintenance at the edge,'' in \emph{2021 31st International Conference on Field-Programmable Logic and Applications (FPL)}.\hskip 1em plus 0.5em minus 0.4em\relax IEEE, 2021, pp. 1--8.

\bibitem{Abhinav2008}
A.~Saxena, K.~Goebel, D.~Simon, and N.~Eklund, ``Damage propagation modeling for aircraft engine run-to-failure simulation,'' \emph{International Conference on Prognostics and Health Management}, 10 2008.

\bibitem{Jinyeong2016}
J.~Moon, P.~Lindahl, J.~Donnal, S.~Leeb, L.~Zachar, L.~Cotta, and C.~Schantz, ``A nonintrusive magnetically self-powered vibration sensor for automated condition monitoring of electromechanical machines,'' in \emph{2016 IEEE AUTOTESTCON}, 09 2016, pp. 1--7.

\bibitem{BoonsLowpoweronlinemachinemonitoringat2021}
B.~Boons, M.~Verhelst, and P.~Karsmakers, ``Low power on-line machine monitoring at the edge,'' in \emph{2021 International Conference on Applied Artificial Intelligence (ICAPAI)}.\hskip 1em plus 0.5em minus 0.4em\relax IEEE, 2021, pp. 1--8.

\bibitem{AsadACurrentSpectrumBasedAlgorithmforF2023}
B.~Asad, H.~A. Raja, T.~Vaimann, A.~Kallaste, R.~Pomarnacki, and K.~{van Hyunh}, ``A current spectrum-based algorithm for fault detection of electrical machines using low-power data acquisition devices,'' \emph{Electronics}, vol.~12, no.~7, p. 1746, 2023.

\bibitem{CN115941729A}
\BIBentryALTinterwordspacing
``\BIBforeignlanguage{en}{Tinyml-based ultra-low power consumption high-reliability end-side predictive maintenance system}.'' [Online]. Available: \url{https://worldwide.espacenet.com/patent/search/family/086698360/publication/CN115941729A?q=pn\%3DCN115941729A}
\BIBentrySTDinterwordspacing

\bibitem{RAY20221595}
\BIBentryALTinterwordspacing
P.~P. Ray, ``A review on tinyml: State-of-the-art and prospects,'' \emph{Journal of King Saud University - Computer and Information Sciences}, vol.~34, no.~4, pp. 1595--1623, 2022. [Online]. Available: \url{https://www.sciencedirect.com/science/article/pii/S1319157821003335}
\BIBentrySTDinterwordspacing

\bibitem{CN116184900A}
\BIBentryALTinterwordspacing
``\BIBforeignlanguage{en}{Passive predictive maintenance system based on tinyml and kinetic energy collection technology}.'' [Online]. Available: \url{https://worldwide.espacenet.com/patent/search/family/086446015/publication/CN116184900A?q=pn\%3DCN116184900A}
\BIBentrySTDinterwordspacing

\bibitem{noauthor_grammarly_nodate}
\BIBentryALTinterwordspacing
``{Grammarly}.'' [Online]. Available: \url{https://app.grammarly.com/}
\BIBentrySTDinterwordspacing

\bibitem{noauthor_deepl_nodate}
\BIBentryALTinterwordspacing
``{DeepL} {Write}: {AI}-powered writing companion.'' [Online]. Available: \url{https://www.deepl.com/write}
\BIBentrySTDinterwordspacing

\end{thebibliography}

\end{document}